\documentclass[twoside]{article}

\usepackage[accepted]{aistats2020}
\usepackage[round]{natbib}

\usepackage[utf8]{inputenc} 
\usepackage[T1]{fontenc}    
\usepackage{hyperref}       
\usepackage{url}            
\usepackage{booktabs}       
\usepackage{amsfonts}       
\usepackage{nicefrac}       
\usepackage{microtype}      

\usepackage{graphicx}
\usepackage{amssymb}
\usepackage{amsmath}
\usepackage{amsthm}
\usepackage{mathrsfs}
\usepackage{bbm}
\usepackage{csquotes}
\usepackage{color}
\usepackage{xcolor}
\usepackage[toc,title,page]{appendix}
\usepackage{booktabs}
\usepackage{mathtools}
\usepackage{multirow}
\usepackage{array}
\usepackage{algorithm}
\usepackage{algorithmic}
\usepackage{subfigure}
\usepackage[font=small,labelfont=bf]{caption}

\usepackage{mathtools}
\usepackage{url}
\usepackage{hyperref}
\usepackage{enumitem}
\usepackage{wrapfig}

\definecolor{mydarkblue}{rgb}{0,0,0.95}
\definecolor{amber}{rgb}{1.0, 0.75, 0.0}
\hypersetup{
    colorlinks=true,
    linkcolor=mydarkblue,
    citecolor=mydarkblue,
    filecolor=mydarkblue,
    urlcolor=mydarkblue}

\newcommand{\lS}{\mathcal{S}}
\newcommand{\AAA}{\mathcal{A}}
\newcommand{\N}{\mathbb{N}}
\newcommand{\F}{\mathcal{F}}
\newcommand{\EE}{\mathbb{E}}
\newcommand{\LL}{\mathcal{L}}

\newcommand{\XX}{\mathcal{X}}
\newcommand{\YY}{\mathcal{Y}}

\newcommand{\EEE}{\mathcal{E}}
\newcommand{\RR}{\mathbb{R}}
\newcommand{\DD}{\mathcal{D}}
\newcommand{\TT}{\mathcal{T}}
\newcommand{\NN}{\mathcal{N}}
\newcommand{\diag}[1]{\operatorname{diag}\left( #1 \right)}
\newcommand{\ovec}{\operatorname{vec}}
\newcommand{\norm}[1]{\left\|#1\right\|}
\newcommand{\Tr}{\operatorname{Tr}}
\DeclarePairedDelimiterX{\infdivx}[2]{(}{)}{%
	#1\;\delimsize\|\;#2%
}

\newtheorem{Theorem}{Theorem}[section]

\theoremstyle{definition}

\newtheorem{Definition}[Theorem]{Definition} 

\begin{document}

%
\runningtitle{Large-Batch Stochastic Gradient Descent with Structured Covariance Noise}

%
\runningauthor{Wen, Luk, Gazeau, Zhang, Chan, Ba}

\twocolumn[

\aistatstitle{An Empirical Study of Large-Batch Stochastic Gradient Descent \\with Structured Covariance Noise}
\aistatsauthor{Yeming Wen$^{*,1,2}$, Kevin Luk$^{*,3}$, Maxime Gazeau$^{*,3}$, Guodong Zhang$^{1,2}$, Harris Chan$^{1,2}$, Jimmy Ba$^{1,2}$}

\aistatsaddress{$^1$ University of Toronto, $^2$ Vector Institute, $^3$ Borealis AI}]

\begin{abstract}
The choice of batch-size in a stochastic optimization algorithm plays a substantial role for both optimization and generalization. Increasing the batch-size used typically improves optimization but degrades generalization. To address the problem of improving generalization while maintaining optimal convergence in large-batch training, we propose to add covariance noise to the gradients. We demonstrate that the learning performance of our method is more accurately captured by the structure of the covariance matrix of the noise rather than by the variance of gradients. Moreover, over the convex-quadratic, we prove in theory that it can be characterized by the Frobenius norm of the noise matrix. Our empirical studies with standard deep learning model-architectures and datasets shows that our method not only improves generalization performance in large-batch training, but furthermore, does so in a way where the optimization performance remains desirable and the training duration is not elongated.
\end{abstract}

\begin{figure}[t]
\centering     
\vspace{-0.1cm}
\begin{center}
\subfigure{\label{subfig:toy_sgd}\includegraphics[height=1.33in,width=0.54\columnwidth]{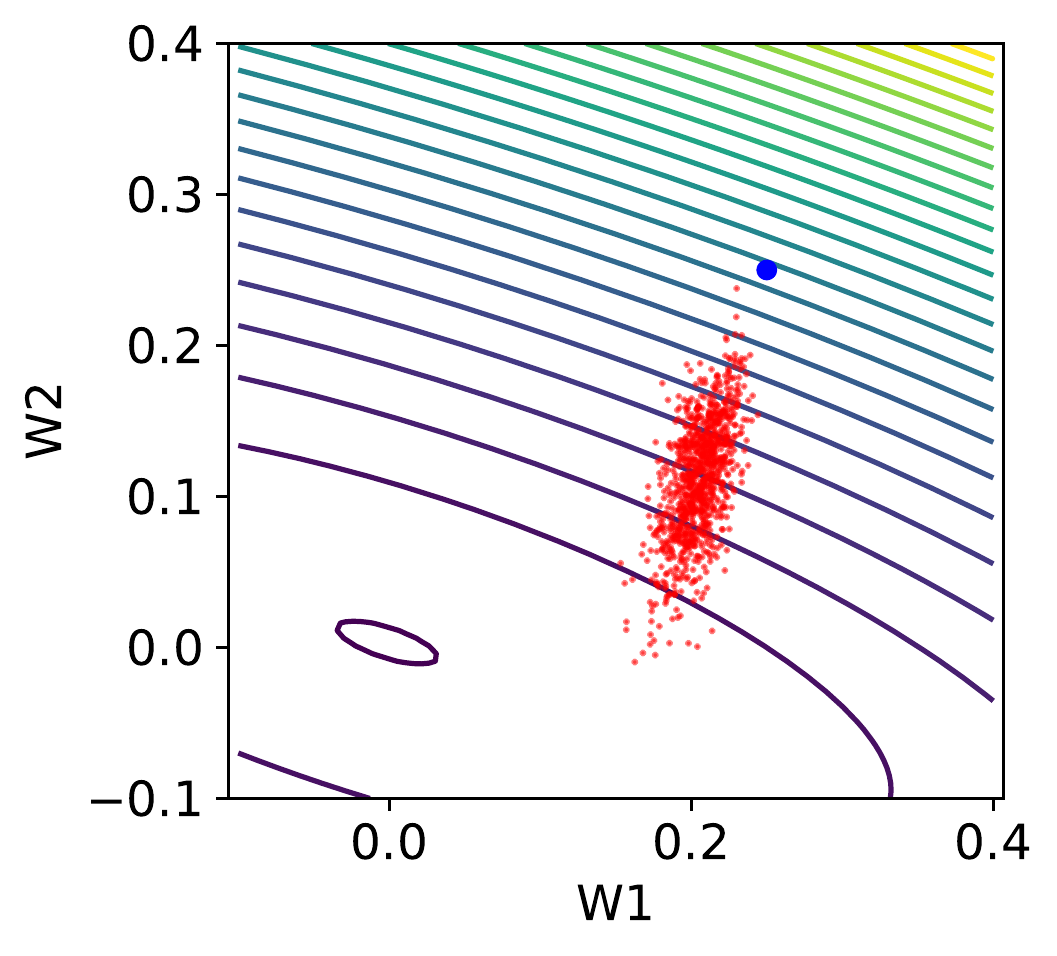}}
\subfigure{\label{subfig:toy_iso}\includegraphics[height=1.33in,width=0.44\columnwidth]{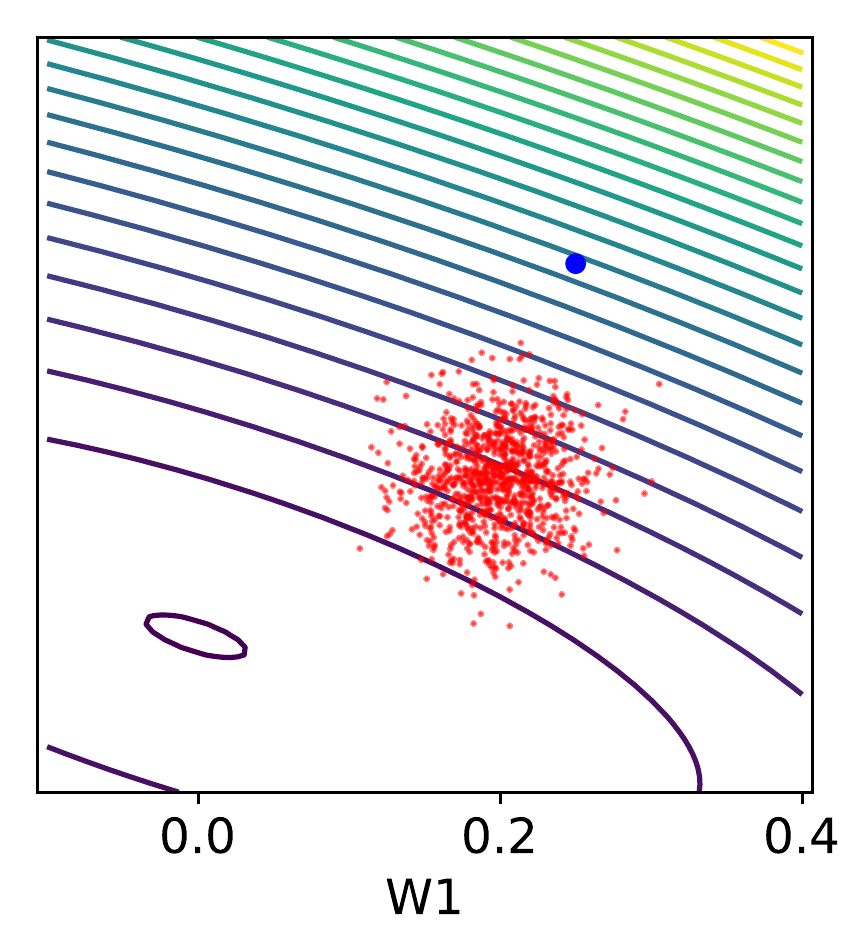}}
\end{center}
\vspace{-0.8cm}
\begin{center}
\subfigure{\label{subfig:toy_fullfisher}\includegraphics[height=1.33in,width=0.54\columnwidth]{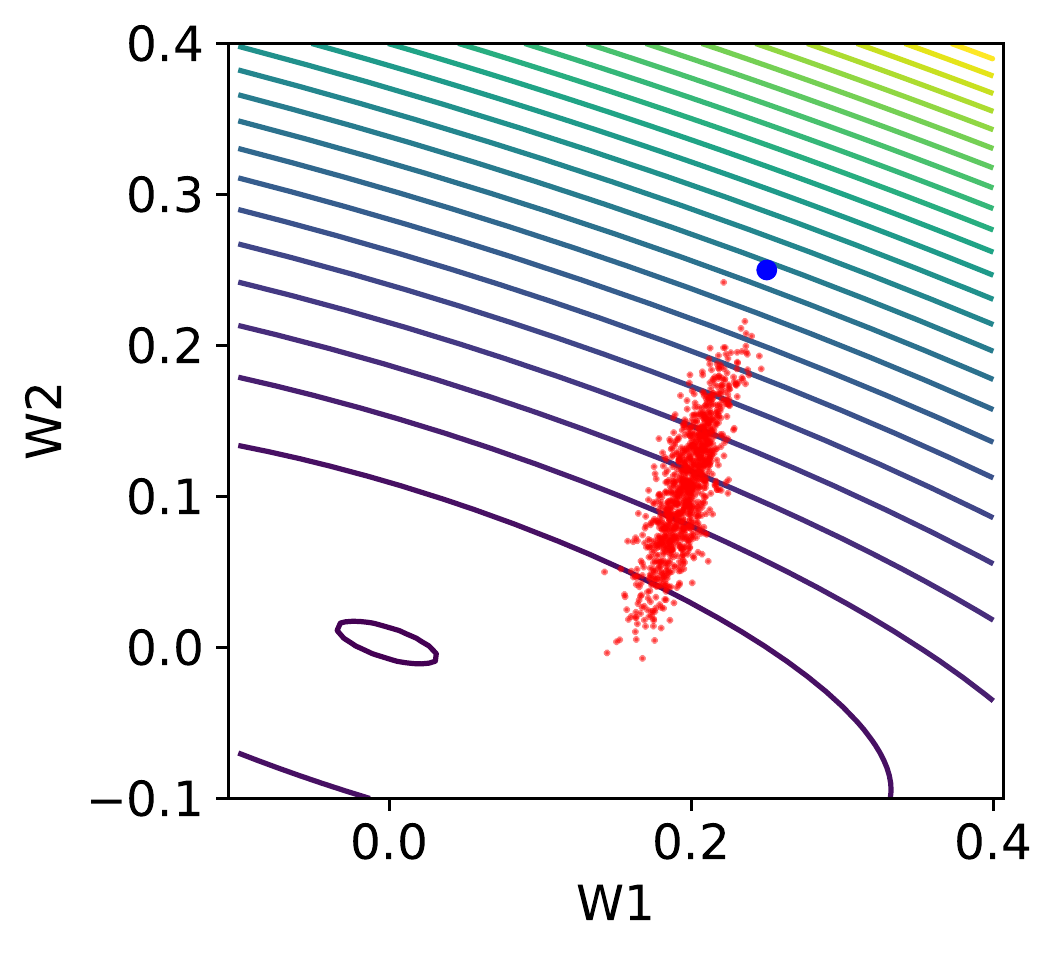}}
\subfigure{\label{subfig:toy_diagfisher}\includegraphics[height=1.33in,width=0.44\columnwidth]{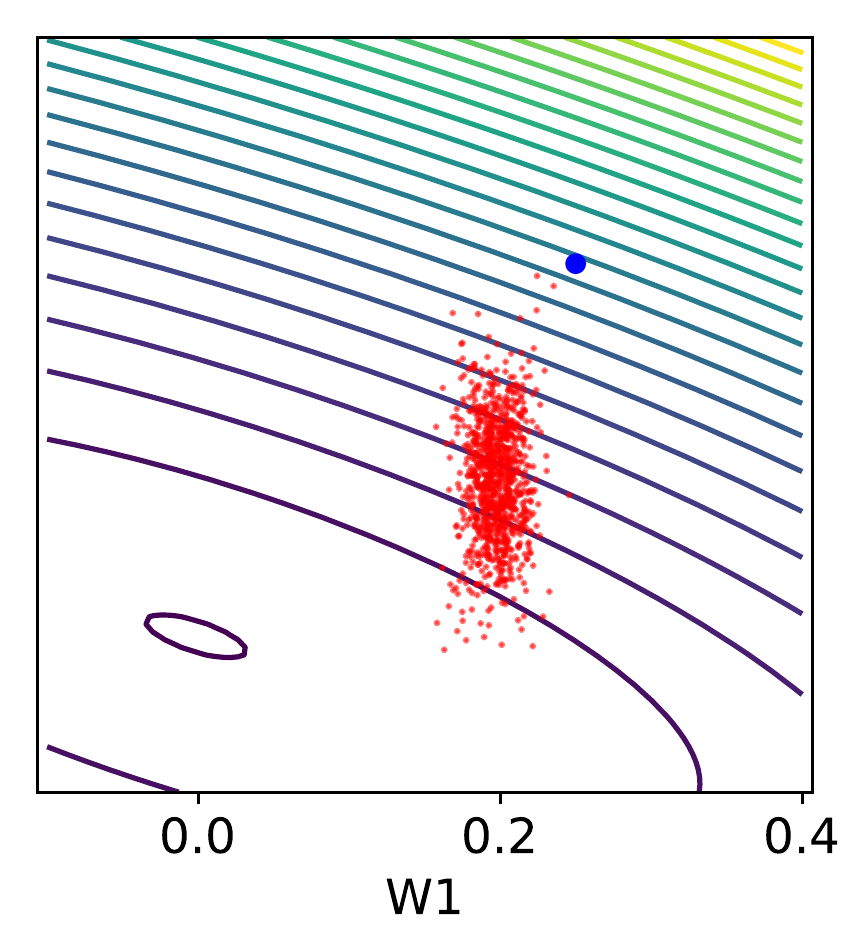}}
\end{center}
\vspace{-0.3cm}
\caption{Noise structure in a simple two-dimensional regression problem. {\bf Top-left}: One-step SGD update. {\bf Top-right}: One-step SGD update with isotropic Gaussian ($\sigma = 0.1$) noise. {\bf Bottom-left}: One-step SGD update with full Fisher noise. {\bf Bottom-right}: One-step SGD update with diagonal Fisher noise. The full Fisher noise almost recovers the SGD noise. Observe that the full Fisher noise direction is perpendicular to the contours of the loss surface. Moreover, full Fisher exhibits slower convergence than diagonal Fisher; we refer to Section~\ref{sec:convex_quadratic} for a more detailed analysis.\label{fig:toy_regression}}
\vspace{-0.3cm}
\end{figure}

\section{Introduction}
\label{sec:introduction}

From a strictly mathematical perspective, training neural networks is a high-dimensional non-convex optimization problem and the dynamics of the training process is incredibly complicated. Despite this, Stochastic Gradient Descent (SGD) and its variants have proven to be extremely effective for training neural networks in practice. Much of the recent successes of deep learning in application tasks such as image recognition~\citep{he2016deep}, speech recognition~\citep{amodei2016deep}, natural language processing~\citep{wu2016google} and game playing~\citep{mnih2015human} can be seen as testaments to the effectiveness of SGD. 

The choice of a batch-size plays an important role in the learning behavior of SGD. Taking larger batch-sizes ensures better gradient estimation which typically leads to faster training convergence. However, there is a tradeoff from the viewpoint of generalization; the intrinsic noise stemming from mini-batch gradients provides regularization effects~\citep{chaudhari2017stochastic,smith2017understanding} and by increasing batch-sizes, we lose such generalization benefits. It is then an interesting question to ask whether large-batch can be engineered in a way such that generalization significantly improves but at the same time not sacrificing too much the training convergence. This is exactly the central objective of our paper.

To address this question, we propose to add a noise term whose covariance structure is given by the diagonal Fisher matrix to the large-batch gradient updates. We discuss the motivations underlying our approach. Under the standard log-likelihood loss assumption, the difference of large-batch gradients and small-batch gradients can be modeled as a Fisher noise. We can expect that adding this noise directly to large-batch gradients will yield small-batch performance. While this may resolve generalization issues associated with large-batch training~\citep{keskar2016large,hoffer2017train}, the resulting convergence performance is undesirable. To attain our ultimate goal of designing a method which enjoys desirable optimization and generalization performance simultaneously, we reduce the noise level by changing the covariance structure from full Fisher to diagonal Fisher.

Variance is commonly regarded as a criteria of optimization performance. However, for our proposed method in this paper, studying the gradient variance is not sufficient in deducing any information on the optimization behavior. Rather, it is the structure of the covariance matrix of the noise which plays a critical role. For large-batch training with diagonal Fisher, we find that despite having a high gradient variance, it still attains an ideal optimization performance.

\textbf{Outline and Summary of Main Contributions.}
We begin in Section~\ref{sec:preliminaries} by introducing the basic framework and necessary definitions. We consider different covariance structures for large-batch training in Section~\ref{subsec:motivations} and then propose the choice of diagonal Fisher. Sections~\ref{sec:initexperiments} and~\ref{sec:convex_quadratic} constitute the central contributions of the paper. The primary takeaways are:

\begin{itemize}
    \item Gradient variance is not an accurate indicator of optimization behavior. In Fig.~\ref{fig:train_loss}, we find empirically that the convergence of large-batch with diagonal Fisher is much faster than that of large-batch with full Fisher and small-batch. However, in Fig.~\ref{fig:marginal_variance}, we find that all these regimes share roughly the same average gradient variance.
    \item The main theoretical contribution is Theorem~\ref{convergence analysis theorem}. We show over the convex quadratic setting, the convergence can be characterized by the Frobenius norm of the noise covariance matrix. In Fig.~\ref{fig:frob_norm}, we show empirically that this carries over to the non-convex deep learning context.
\end{itemize}

In Section~\ref{sec:moreexperiments}, we apply our methodology to address the~\enquote{generalization gap} problem. We show that within the same number of training epochs, large-batch with diagonal Fisher can attain generalization performance roughly comparable to that of small-batch. Related works are discussed in Section~\ref{sec:related works} and we close the paper in Section~\ref{sec:conclusion}.

\section{Preliminaries and Approach}
\label{sec:preliminaries}

\subsection{Preliminary Background}

\textbf{Excess Risk Decomposition.} We work in the standard framework of supervised learning. Let $\DD$ be the unknown joint probability distribution over the data domain $\XX\times\YY$ where $\XX$ is the input space and $\YY$ is the target space. We have a training set $\lS=\{(x_1,y_1),\dots,(x_N,y_N)\}$ of $N$ input-target samples drawn i.i.d. from $\DD$. The family of classifiers of interest to us are neural network outputs $f(x_i,\theta)$, where $\theta\in\RR^d$ are parameters of the network. Let $\ell : \YY \times \YY \to \mathbb{R}$ 
be the loss function measuring the disagreement between outputs $f(x_i,\theta)$ and targets $y_i$. For convenience, we use the notation $\ell_i(\theta)$ to denote $\ell(f(x_i,\theta),y_i)$. The expected risk and empirical risk functions are defined to be
\[
\mathscr{L}(\theta):=\EE_{(x,y)\sim\DD}[\ell(f(x,\theta),y)],\ \LL(\theta):= \frac{1}{N}\sum^N_{i=1}\ell_i(\theta). 
\]
The standard technique to analyze the interplay between optimization and generalization in statistical learning theory is through excess risk decomposition. The excess risk, after $k$ iterations, is defined as:
\[
\Delta_k := \mathscr{L}(\theta_k) - \inf_{\theta \in \RR^d}\mathscr{L}(\theta).
\]

From~\citet{bottou2008tradeoffs, CJY:18}, the expected excess risk can be upper-bounded by
\begin{equation} \label{eq:opt-gen upper bound}
\EE_{\lS}[\Delta_k]\leq\underbrace{\EE_{\lS}[\mathscr{L}(\theta_k)-\LL(\theta_k)]}_{\EEE_{\mathrm{gen}}}+\underbrace{\EE_{\lS}[\LL(\theta_k)-\LL(\theta^*)]}_{\EEE_{\mathrm{opt}}}.
\end{equation}
where $\theta^*=\mathrm{argmin}_\theta\LL(\theta)$ here is the empirical risk minimizer. The terms $\EEE_{\mathrm{gen}}$ and $\EEE_{\mathrm{opt}}$ are the expected generalization and optimization errors respectively. It is often the case that optimization algorithms are studied from one perspective: either optimization or generalization. The decomposition in Eqn.~\ref{eq:opt-gen upper bound} indicates that both aspects should be analyzed
together~\citep{bottou2008tradeoffs,CJY:18}; since the goal of a good optimization-generalization algorithm in machine learning is to minimize the excess risk in the least amount of iterations.

\subsection{Motivations and Approach}
\label{subsec:motivations}

We begin by formalizing the setup. Let $B_L$ denote large-batch and $M_L=|B_L|$ denote the size of the large-batch. We consider the following modification of large-batch SGD updates
\begin{equation} \label{eq:naive LB update}
\theta_{k+1} = \theta_k - \alpha_k\nabla\LL_{M_L}(\theta_k) + \alpha_k C(\theta_k)\xi_k.
\end{equation}
where $\alpha_k$ is the learning rate, $\xi_k\sim\NN(0,I_d)$ is the multivariate Gaussian distribution with mean zero and identity covariance, and $\nabla\LL_{M_L}(\theta_k)=\frac{1}{M_L}\sum_{i\in B_L}\nabla\ell_i(\theta_k)$ is the large-batch gradient. We can interpret Eqn.~\ref{eq:naive LB update} as modifying large-batch SGD by injecting Gaussian noise with mean zero and covariance $C(\theta_k)C(\theta_k)^{\top}$ to the gradients. The central goal of this paper is to determine a suitable matrix $C(\theta_k)$ such that the excess risk of the algorithm in Eqn.~\ref{eq:naive LB update} is minimized; in more concrete terms, it achieves low optimization and generalization error simultaneously within a reasonable computational budget.

\subsubsection{Intrinsic SGD Noise}
Let $B\subset\lS$ be a mini-batch drawn uniformly and without replacement from $\lS$ and $M=|B|$ be the size of this chosen mini-batch. We can write the SGD update rule here as
\begin{align*}
\theta_{k+1} & = \theta_k - \alpha_k\nabla\LL_M(\theta_k) \\
& = \theta_k - \alpha_k\nabla\LL(\theta_k) + \alpha_k(\underbrace{\nabla\LL(\theta_k)-\nabla\LL_M(\theta_k)}_{\delta_k}) 
\end{align*}
where $\nabla\LL(\theta_k)=\frac{1}{N}\sum^N_{i=1}\nabla\ell_i(\theta_k)$ is the full-batch gradient. The difference $\delta_k=\nabla\LL(\theta_k)-\nabla\LL_M(\theta_k)$ is the intrinsic noise stemming from mini-batch gradients. The covariance of $\delta_k$ is given by
\begin{equation} \label{eq:covariance matrix}
\frac{N-M}{M}\frac{1}{N}\sum^N_{i=1}\left(\nabla\LL(\theta_k)-\nabla\ell_i(\theta_k)\right)\left(\nabla\LL(\theta_k)-\nabla\ell_i(\theta_k)\right)^{\top} 
\end{equation}
This result can be found in~\citet{hu2017diffusion,hoffer2017train}. Moreover, this type of noise has been studied in~\citet{ZWYM:18}. For the purposes of this paper, we assume that the loss is taken to be negative log-likelihood, $\ell_i(\theta_k)=-\log p(y_i|x_i,\theta_k)$ where $p(y|x,\theta)$ is the density function for the model's predictive distribution. Moreover, we assume that the gradient covariance matrix above can be approximated by
\begin{equation} \label{eq:empirical Fisher}
\frac{N-M}{M}\underbrace{\frac{1}{N}\sum_{i=1}^N\nabla \log p(y_i|x_i,\theta_k)\nabla \log p(y_i|x_i,\theta_k)^{\top}}_{F(\theta_k)}, \end{equation}
where $(x_i,y_i)$ are sampled from the empirical data distribution. In the literature, the matrix $F(\theta_k)$ above is often referred to as the empirical Fisher matrix~\citep{martens2014new}. We make this approximation for two reasons. First, computing the full-batch gradient $\nabla\LL(\theta_k)$ at every iteration is not feasible computationally. Secondly, from Fig.~\ref{fig:train_loss}, we find empirically that the training dynamics of a large-batch regime with empirical Fisher is very close to a small-batch regime (which by the above analysis should be captured by large-batch with empirical covariance in Eqn.~\ref{eq:covariance matrix}); suggesting that it is a reasonable assumption to make.

For the remainder of this paper, unless otherwise specified, all mentions of~\enquote{Fisher matrix} or $F(\theta)$ refers to the empirical Fisher. For completeness, we provide explicit expressions of diagonal Fisher for feed-forward and convolutional network architectures in Appendix~\ref{sec:Fisher for different architectures}.

\subsubsection{Naive Choices of Covariance Matrices}
We begin by considering the choice of $C(\theta_k)=0$. In this case, Eqn.~\ref{eq:naive LB update} is just standard large-batch gradient descent. Since large-batches provide better gradient estimation, we can expect better training error per parameter update. 
However, from the perspective of generalization, it has been observed in~\citet{lecun1998gradient,keskar2016large,hoffer2017train} that using larger batch-sizes can lead to a decay in generalization performance of the model.

Now, let $B_S, B_L$ denote small-batch and large-batch, $M_S=|B_S|, M_L=|B_L|$ denote the size of small-batch and large-batch. Consider $C(\theta_k)$ to be
\begin{equation} \label{eq:full Fisher noise choice}
C(\theta_k)=\sqrt{\frac{M_L-M_S}{M_L M_S}}\sqrt{F(\theta_k)}.
\end{equation}
Now, if the intrinsic SGD noise is reasonably approximated as a Gaussian distribution with mean zero and covariance given by $C(\theta_k)$ above, then Eqn.~\ref{eq:naive LB update} with this choice of $C(\theta_k)$ should exhibit similar behavior as small-batch. If this is the case, then we can expect that Eqn.~\ref{eq:naive LB update} exhibits poor convergence.
Indeed, as shown on a 2D convex example in Fig.~\ref{fig:toy_regression}, choosing $C(\theta_k)$ as in Eqn.~\ref{eq:full Fisher noise choice} essentially recovers SGD behavior. Furthermore, on the CIFAR-10 image classification task trained using ResNet-44 in Fig.~\ref{fig:train_loss}, we find that adding this noise significantly worsens the training convergence. Thus, choosing $C(\theta_k)$ as in Eqn.~\ref{eq:full Fisher noise choice} does not satisfy our objective of simultaneously attaining desirable convergence and generalization for large-batch training.

\subsubsection{Using Diagonal Fisher} 
We now propose to take a~\enquote{middle ground} and choose $C(\theta_k)$ to be
\begin{equation} \label{eq:diagonal Fisher matrix choice}
C(\theta_k)=\sqrt{\frac{M_L-M_S}{M_L M_S}}\sqrt{\diag{F(\theta_k)}}.
\end{equation}
A formal statement is given in Algorithm~\ref{alg:diagFnoise}. Changing from full Fisher to diagonal Fisher has important implications for both optimization and generalization behavior. In our empirical analysis in Sections~\ref{sec:initexperiments} and~\ref{sec:moreexperiments}, we show that Algorithm~\ref{eq:opt-gen upper bound} can achieve both desirable convergence and generalization performance within an epoch training budget; which implies that the excess risk is minimized.

With regards to computational complexity, computing diagonal Fisher only introduces minor overhead.~\citet{Goodfellow2015EfficientPG} shows that it can be done at the cost of one forward pass.

\begin{algorithm}[t] 
	\caption{Adding diagonal Fisher noise to large-batch SGD. Differences from standard large-batch SGD are highlighted in \color{blue} blue\color{black}}
	\label{alg:diagFnoise}
	\begin{algorithmic}
		\color{black}
		\REQUIRE Number of iterations $K$, initial step-size $\alpha_0$, large-batch $B_L$ of size $M_L$, small-batch $B_S$ of size $M_S$, initial condition $\theta_0\in\RR^d$\FOR{$k=1$ to $K$}
		\STATE \color{blue}$\xi_k \sim \mathcal{N}(0, I_d)$ 
		\STATE $\epsilon_k = \alpha_k\sqrt{\frac{M_L-M_S}{M_LM_S}}\sqrt{\diag{F(\theta_k)}}\xi_k$ 
		\STATE \color{black} $\theta_{k+1} = \theta_{k} - \alpha_k \nabla \LL_{M_L}(\theta_{k}) \color{blue} + \epsilon_k \color{black}$  
		\ENDFOR
	\end{algorithmic}
\end{algorithm}

\section{Variance and Optimization}
\label{sec:initexperiments}

The objective of this section is to examine the optimization performance of the four regimes: large-batch with $C(\theta_k)$ equal to $0$ (standard large-batch), large-batch with $C(\theta_k)$ equal to diagonal Fisher as in Eqn.~\ref{eq:diagonal Fisher matrix choice}, large-batch with $C(\theta_k)$ equal to full Fisher as in Eqn.~\ref{eq:full Fisher noise choice} and small-batch. In the experimentation, we fix large-batch to be 4096 and small-batch to be 128. For conciseness, we set forth the notation {\bf LB} and {\bf SB} for large-batch and small-batch respectively.

\begin{figure}
\begin{center}
\includegraphics[width=0.75\columnwidth]{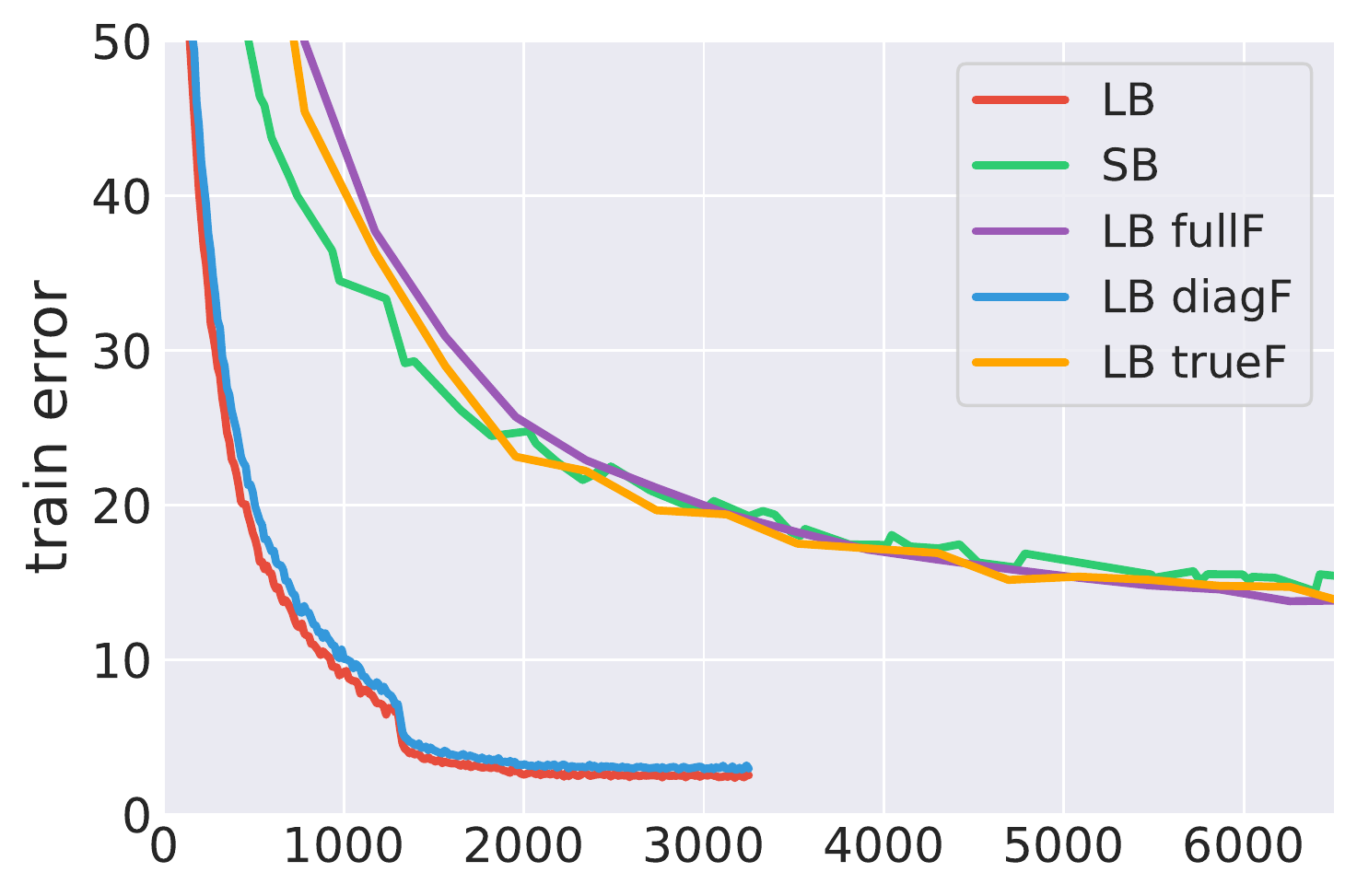}
\caption{Training error per parameter update for {\bf SB}, {\bf LB}, {\bf LB} with full Fisher and {\bf LB} with diagonal Fisher. \label{fig:train_loss}}
\end{center}
\vspace{-10mm}
\end{figure}

In Fig.~\ref{fig:train_loss}, we compare the training error (measured per parameter update) of ResNet44 (CIFAR-10) of the four regimes. The same learning rate is used across all four regimes (better training error can be obtained if we tune the learning rate further for the {\bf LB} regimes). We find that {\bf LB} with diagonal Fisher trains much faster than {\bf LB} with full Fisher and {\bf SB}. In contrast, {\bf LB} with diagonal Fisher attains a convergence similar to {\bf LB}, demonstrating that adding this particular form of noise does not hamper much the optimization performance. 

\begin{figure}
\begin{center}
\includegraphics[width=0.75\columnwidth]{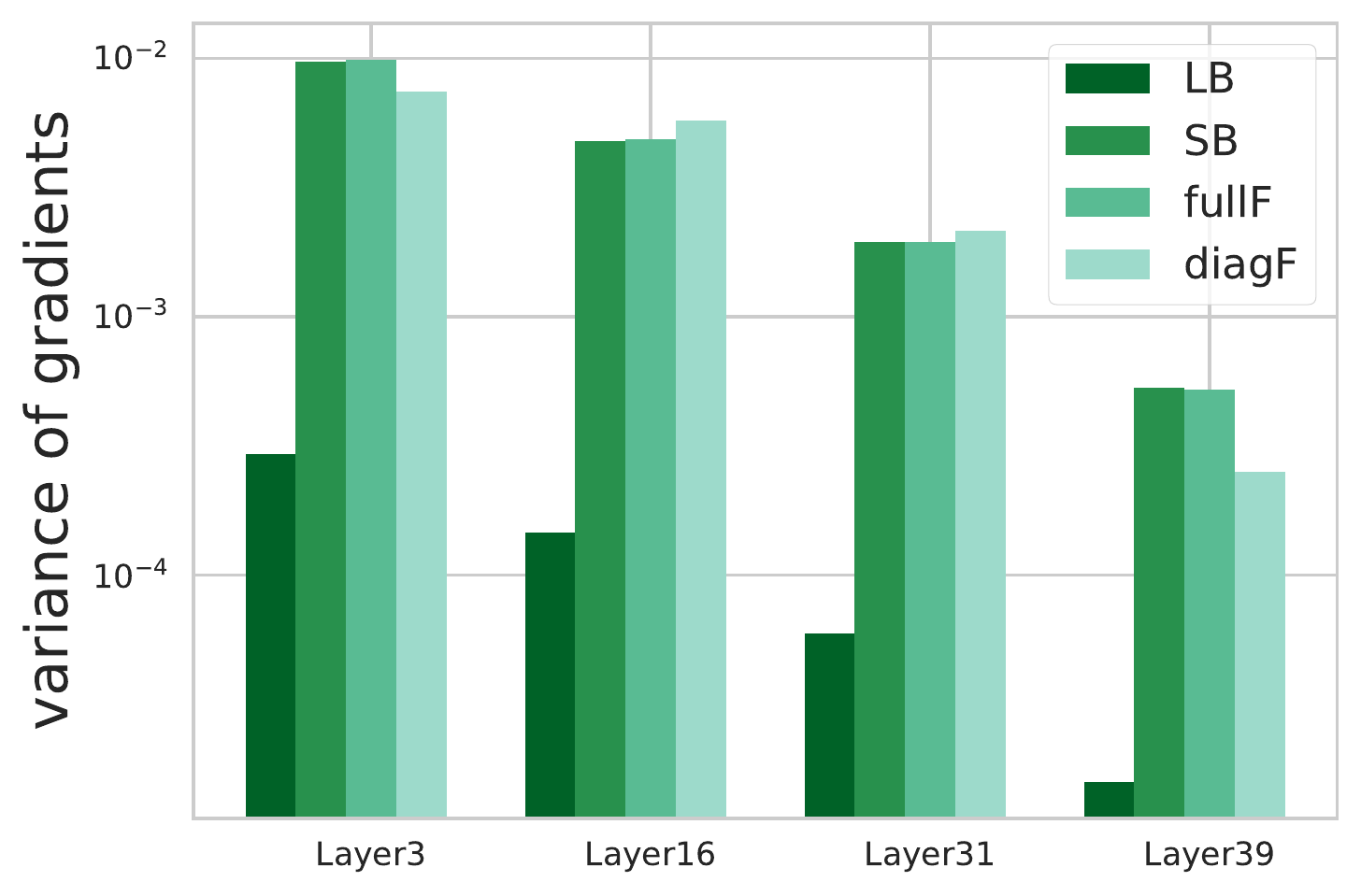}
\caption{Average variance of gradients for {\bf LB}, {\bf SB}, {\bf LB} with full Fisher and {\bf LB} with diagonal Fisher. \label{fig:marginal_variance}}
\end{center}
\vspace{-10mm}
\end{figure}

We now analyze the gradient variance of each of the four regimes. We define gradient variance here to be the trace of the covariance matrix of the gradients. In the experiment depicted in Fig.~\ref{fig:marginal_variance}, we provide an estimation of the variance of gradients of the four regimes. The experiment is performed as follows: we freeze a partially-trained network and compute Monte-Carlo estimates of gradient variance with respect to each parameter over different mini-batches. This variance is then averaged over the parameters within each layer. 

In Fig.~\ref{fig:marginal_variance}, we find that {\bf LB} with diagonal Fisher, {\bf LB} with full Fisher, and {\bf SB} all share roughly the same gradient variance meanwhile {\bf LB} has a much lower one. 

However, the optimization behaviors are completely different. These experiments show that the number of iterations needed to reach a small optimization error is not purely determined by the gradient variance.
Many recent works have suggested to add isotropic noise to the gradient dynamics to escape from saddle point or local minima~\citep{2015arXiv150302101G, 2017arXiv170300887J}. 
We believe that the number of iterations needed to escape saddle points is determined by the nature of the noise. A similar observation has been made in~\citet{ZWYM:18} where the authors studied the effects of how using Fisher covariance noise relates to the efficiency of escaping from local minima and compare it to isotropic noise.

In the next section, we analyze the efficiency of the algorithm in Eqn.~\ref{eq:naive LB update} for a convex quadratic when the covariance matrix is given by the (exact true) Fisher matrix. We prove that it requires less iterations to reach the global optimum compared to isotropic noise. We iterate that this choice of diffusion matrix is completely specific to the convex quadratic example.

\section{Convex Quadratic Example}
\label{sec:convex_quadratic}

\subsection{Motivations}
In this section, we analyze the optimization behavior of our proposed algorithm for the convex quadratic model. Analyzing the convex quadratic model serves as a good proxy to understand the complex dynamics of neural network training. Although the optimization landscape of neural networks is non-convex, using such a model marginalizes away inessential features and enables us to tractably analyze optimization phenomena of neural networks. There is ample evidence suggesting this: the recent work of~\citet{zhang2019algorithmic} uses the convex quadratic model to accurately predict critical batch-sizes for commonly used optimizers in neural networks. Short-horizon bias phenomena of optimized learning rates were studied in~\citet{Wu2018UnderstandingSB} for the convex quadratic and their theoretical insights were successfully translated to neural networks. Furthermore, a convex quadratic objective can always be obtained by first linearizing around a given parameter vector and then taking a second-order Taylor approximation. Recent empirical work~\citep{Lee2019WideNN} have demonstrated that linearized approximations do indeed match the training phenomena of large yet realistic networks.

Therefore, approximating the loss surface of a neural network with a convex quadratic has proven to be a fertile~\enquote{testing ground} when introducing new methodologies in deep learning. Analyzing the toy quadratic problem has led to important insights; for example, in learning rate scheduling~\citep{schaul2013no} and formulating SGD as approximate Bayesian inference~\citep{mandt2017stochastic}. 

\subsection{Analysis}
\begin{table*}[!ht]
\caption{Number of iterations needed for the algorithm in Eqn.~\ref{eq:naive LB update} to reach an $\epsilon$ error for a strongly-convex quadratic loss function. Since $\norm{\diag{A}}_{\mathrm{Frob}}^2$ is smaller than $\norm{A}_{\mathrm{Frob}}^2$, less iterations are required when choosing $C = \sqrt{\diag{A}}$.}
\label{tab:summary}
\begin{center}
\begin{small}
\begin{sc}
\begin{tabular}{lcr}
\toprule
covariance matrix $C$ & $\sqrt{A}$ & $\sqrt{\diag{A}}$ \\
\midrule
steps $k$ to reach $\epsilon$ error &  $\norm{A}_{\mathrm{Frob}}^2 /\epsilon$  & $\norm{\diag{A}}_{\mathrm{Frob}}^2 /\epsilon$  \\
\bottomrule
\end{tabular}
\end{sc}
\end{small}
\end{center}
\vskip -0.1in
\end{table*}
For strongly-convex objective functions and diminishing step-sizes, the expected optimality gap is bounded in terms of the second-order moment of the gradients~\citep{bottou2018optimization}. 
However, in practice, different algorithms having the same gradient moments might 
not need the same number of iterations to converge to the minimum.

Consider the loss function 
\[
\LL(\theta) = \frac{1}{2}\theta^{\top}A\theta,
\]
where $A$ is a symmetric, positive-definite matrix. We focus on the algorithm in Eqn.~\ref{eq:naive LB update} and consider a constant $d\times d$ covariance matrix $C$. The following theorem, adapted from~\citet{bottou2018optimization}, analyzes the convergence of this optimization method. 
The proof is relegated to Section~\ref{sec:proof of convergence theorem} of Appendix.

\begin{Theorem} \label{convergence analysis theorem}
Let $\lambda_{\max}$ and $\lambda_{\min}$ denote the maximum and minimum eigenvalue of $A$ respectively. For a chosen $\alpha_0\leq \lambda_{\max}^{-1}$, suppose that we run the algorithm in Eqn.~\ref{eq:naive LB update} according to the decaying step-size sequence
\[
\alpha_k = \frac{2}{(k+\gamma)\lambda_{\min}},
\]
for all $k\in\N_{>0}$ and where $\gamma$ is chosen such that $\alpha_k \leq \alpha_0$. Then for all $k\in\N$, 
\[
\EE[\LL(\theta_k)] \leq \frac{\nu}{k+\gamma}
\]
where
\[
\nu = \max \left( \frac{2 \Tr(C^{\top}AC)}{
            \lambda^2_{\min}} , \gamma\LL(\theta_0) \right).
\]

\end{Theorem}

\begin{figure}
\begin{center}
\includegraphics[width=0.85\columnwidth,height=0.24\textheight]{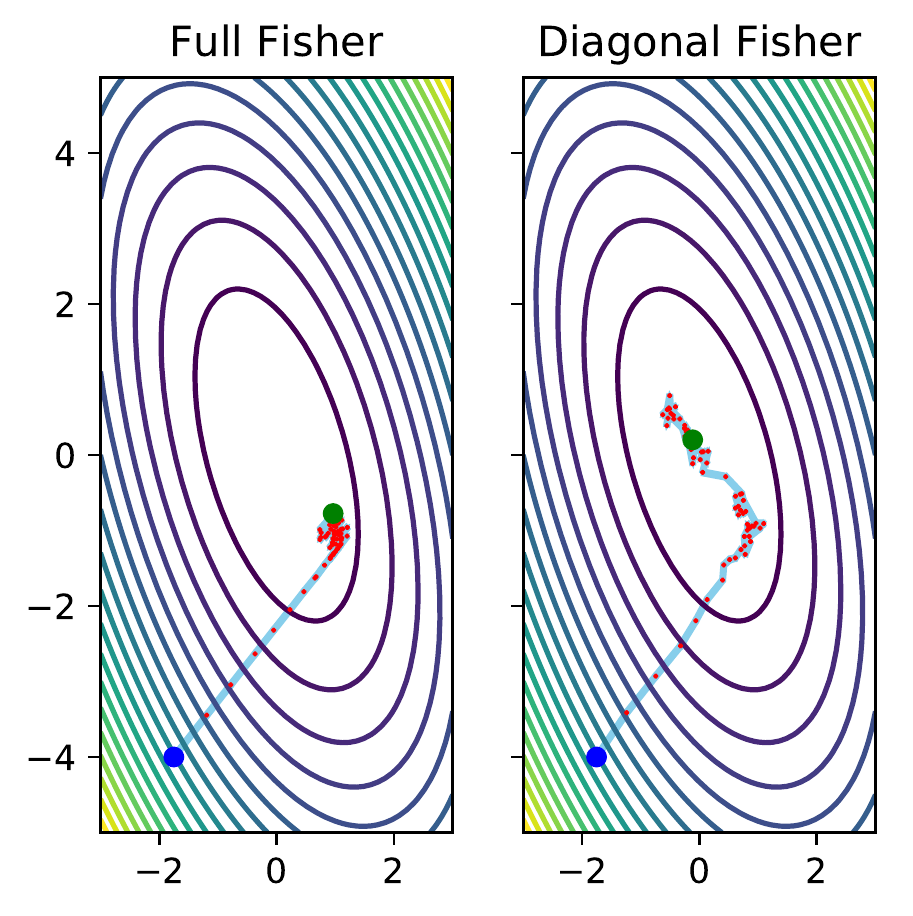}
\caption{Trajectory using full Fisher versus diagonal Fisher noise for the algorithm in Eqn.~\ref{eq:naive LB update} used to minimize a two-dimensional quadratic function. Blue dot indicates the initial parameter value and the green dot shows the final parameter value. We used a learning rate of 0.1 for 500 iterations (plotting every 10 iterations). Observe that adding diagonal Fisher to the gradient achieves faster convergence than full Fisher.\label{fig:toy_trajectory}}
\end{center}
\vspace{-0.3cm}
\end{figure}

We make a couple of observations concerning this bound. First, the convergence rate is optimal when $C=0$ which is expected. In this case, there is no noise and hence we obtain no regularization benefits which leads to poor generalization. A more formal discussion is given at the end of Section~\ref{sec:covariance-generalization} in the Supplementary Material where if we employ a scaling factor $C_\lambda:=\lambda C$; as $\lambda\to0$, the expected generalization error becomes worse. 

The second observation is that the term of importance in this theorem is $\Tr(C^{\top}AC)$. While the overall convergence rate of the algorithm is $O(1/k)$, the discrepancy in convergence performance for different choices of the matrix $C$ rests entirely on this term. The number of iterations for the algorithm in Eqn.~\ref{eq:naive LB update} to reach the unique minimum depends entirely on $\Tr(C^{\top}AC)$ and not on the second-order moment.

We analyze two specific cases which are relevant for us: the first case where $C$ is square-root of $A$, $C=\sqrt{A}$, and the second case where $C$ is the square-root of the diagonal of $A$, $C=\sqrt{\diag{A}}$. 
The second-order moment of the noise perturbation is the same for both and is given by
\begin{equation} 
\begin{aligned}
\EE_{\xi}[\norm{C \xi_k }^2] = \Tr(C^{\top}C) = \Tr(A).
\end{aligned}
\end{equation}
However, it is different for $\Tr(C^{\top}AC)$; in the case of $C=\sqrt{A}$, we get
\[
\Tr(C^{\top}AC)=\Tr(A^2)=\norm{A}_{\mathrm{Frob}}^2,
\]
and for the case of $C=\sqrt{\diag{A}}$,
\[
\Tr(C^{\top}AC) = \Tr(\diag{A}^2) = \norm{\diag{A}}_{\mathrm{Frob}}^2.
\]
Thus, the difference in training performance between the two cases can be measured by the difference of their respective Frobenius norms and
less number of iterations are needed with the choice of $\sqrt{\diag{A}}$. 
This suggests that the off-diagonal elements of $A$ play a role in the optimization performance. In Fig.~\ref{fig:toy_trajectory}, we provide a visualization of the difference between $C=\sqrt{A}$ and $C=\sqrt{\diag{A}}$ over a two-dimensional quadratic function.

We summarize our observation in Table~\ref{tab:summary}: different choices of covariance matrix $C$ impacts the number of iterations required to reach an $\epsilon$ error.

{\bf Limitations:}
We mention a couple of limitations of the current analysis. In the previous analysis, the matrix $A$ above is the Hessian which is also, in this specific setup, equal to the (exact true) Fisher 
defined as the expectation of the outer product of log-likelihood gradients,
\begin{equation} \label{eq:true Fisher}
\EE_{P_x , P_{y|x} }[\nabla \log p(y|x,\theta)\nabla\log p(y|x,\theta)^{\top}]
\end{equation}
The expectation here is taken with respect to the data distribution $P_x$ for inputs $x$ and the model's predictive distribution $P_{y|x}$ for targets $y$. For much of this paper, we have been working with the empirical Fisher instead. We believe this to be a reasonable approximation; in Fig.~\ref{fig:train_loss}, we find that the training curves for {\bf LB} + full Fisher and {\bf LB} + true Fisher are almost identical. Furthermore, if we assume that
the implicit conditional distribution over the network’s output is close to the conditional distribution of targets from the training distribution, then the covariance of the gradients closely matches the Hessian~\citep{martens2014new}. The recent work of~\citet{zhang2019algorithmic} shows that this relationship indeed holds tightly in empirical settings. 

Secondly, we have focused solely on the optimization performance of our proposed method. While it would be desirable to accompany this with a complete theoretical analysis of generalization performance, this is beyond the current scope of the paper. However, in Appendix~\ref{sec:covariance-generalization}, we use the framework of uniform stability to provide some theoretical insights on how different choices of the covariance matrix impact generalization.

\section{Experiments}
\label{sec:moreexperiments}

\subsection{Experimentation Details}

\textbf{Batch Normalization.} For all experiments involving {\bf LB}, we adopt Ghost Batch Normalization (GBN)~\citep{hoffer2017train} and hence {\bf LB} throughout stands for {\bf LB} with Ghost Batch Normalization. This allows a fair comparison between {\bf LB} and {\bf SB}, as it ensures that batch normalization statistics are computed on the same number of training examples. Using standard batch normalization for large batches can lead to degradation in model quality~\citep{hoffer2017train}.

\textbf{Models and Datasets.} The network architectures we use are fully-connected networks, shallow convolutional networks (LeNet~\citep{lecun1998gradient}, AlexNet~\citep{krizhevsky2012imagenet}), and deep convolutional networks (VGG16~\citep{simonyan2014very}, ResNet44~\citep{he2016deep}, ResNet44x2 (the number of filters are doubled)). These models are evaluated on the standard deep-learning datasets: MNIST, Fashion-MNIST~\citep{lecun1998gradient,xiao2017fashion}, CIFAR-10 and CIFAR-100~\citep{krizhevsky2009learning}.

\subsection{Frobenius Norm} 

Over the convex quadratic setting in Section~\ref{sec:convex_quadratic}, Theorem~\ref{convergence analysis theorem} tells us that the number of iterations to reach optimum is characterized by the Frobenius norm. Hence, the optimization difference between using large-batch with diagonal Fisher versus full Fisher lies in the difference of their respective Frobenius norms.

We now give an empirical verification of this phenomena in the non-convex setting of deep neural networks. We compute the Frobenius norms during the training of the ResNet44 network on CIFAR-10. Fig.~\ref{fig:frob_norm} shows that the full Fisher matrix has much larger Frobenius norm than the diagonal Fisher matrix, which suggests that using diagonal Fisher noise should have faster convergence than full Fisher noise in the deep neural network setting. Indeed, Fig.~\ref{fig:train_loss} shows that {\bf LB} with full Fisher converges at the same rate as {\bf SB} whereas {\bf LB} with diagonal Fisher converges much faster; and in fact, roughly the same as {\bf LB}. This indicates that adding diagonal Fisher noise to {\bf LB} does not degrade the optimization performance of {\bf LB}.

\vspace{-1mm}
\subsection{Maximum Eigenvalue of Hessian} 

While the relationship between the curvature of the loss surface landscape and generalization is not completely explicit, numerous works have suggested that the maximum eigenvalue of the Hessian is possibly correlated with generalization performance~\citep{keskar2016large,chaudhari2016entropy,chaudhari2017stochastic,yoshida2017spectral,xing2018walk}. In this line of research, the magnitudes of eigenvalues of the Hessian may be interpreted as a heuristic measure for generalization; the smaller the magnitude the better the model generalizes. To situate our method with this philosophy, we compute the maximum eigenvalue of the Hessian of the final model for the following three regimes: {\bf SB}, {\bf LB}, and {\bf LB} with diagonal Fisher. 

\begin{figure}
\centering     
\subfigure[Frobenius Norm]{\label{fig:frob_norm}\includegraphics[width=0.48\columnwidth]{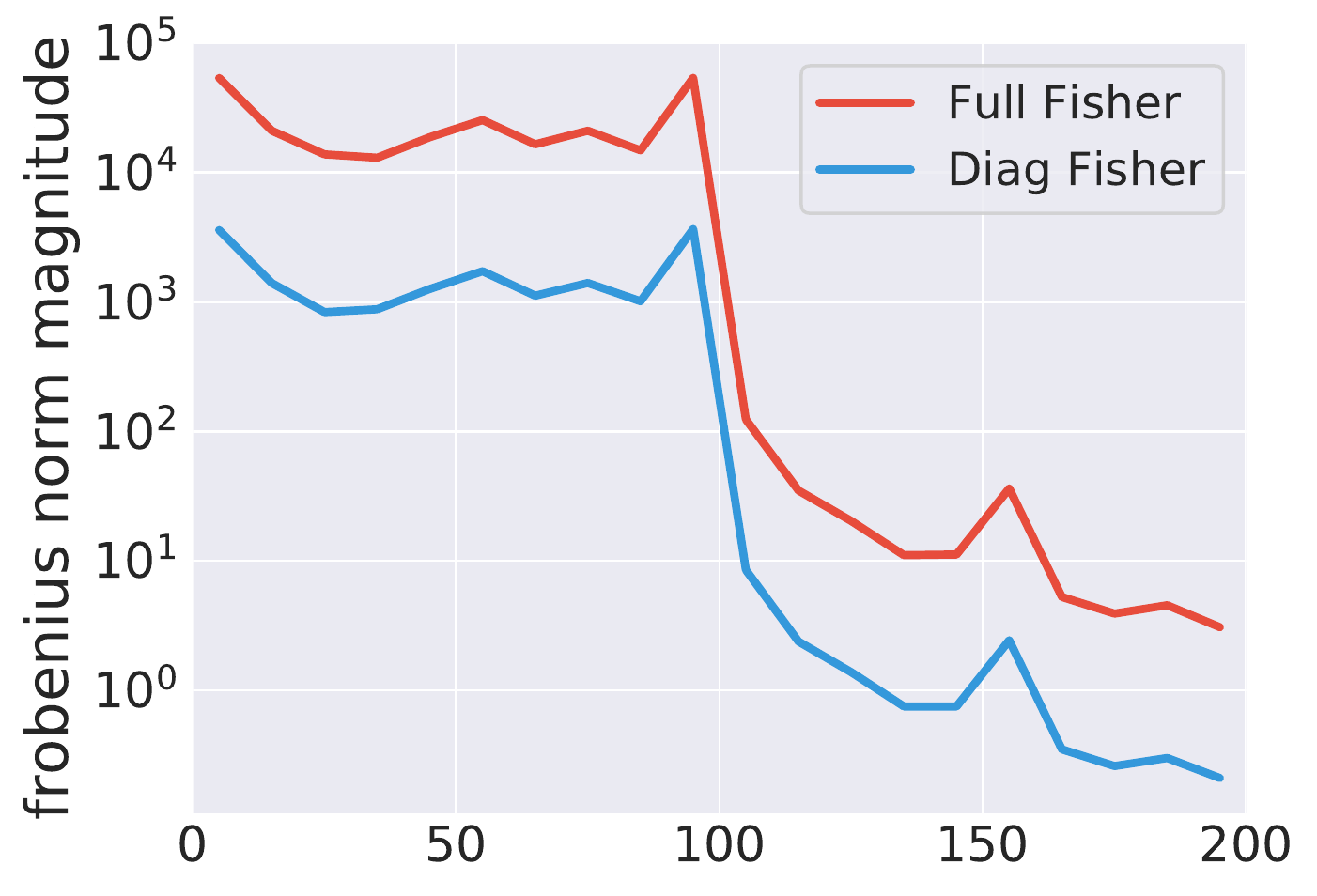}}
\subfigure[Maximum Eigenvalue]{\label{fig:spectral}\includegraphics[width=0.48\columnwidth]{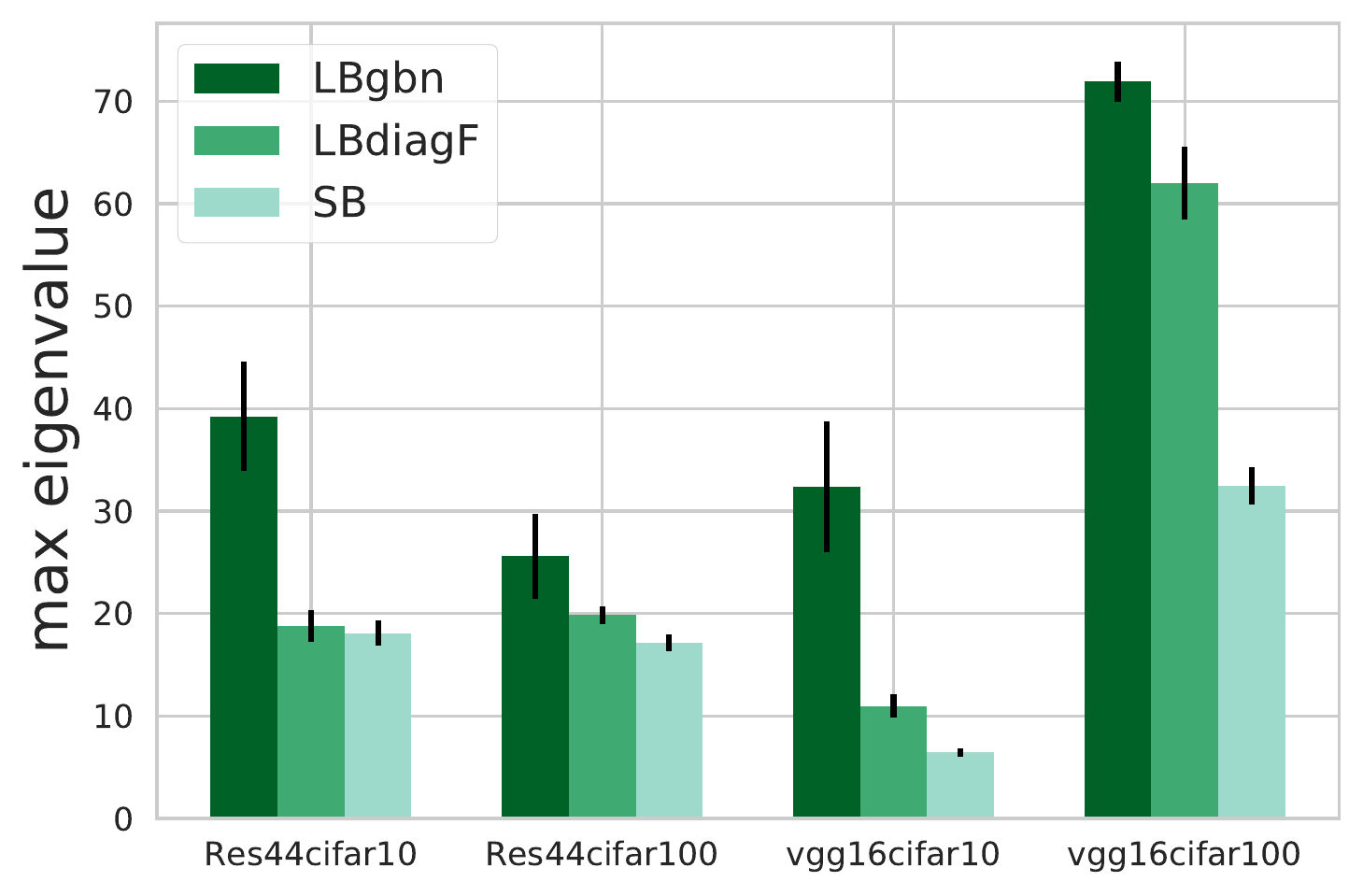}}
\caption{{\bf a)} Frobenius norms of full Fisher and diagonal Fisher along the training trajectory. The model is trained on ResNet44 with CIFAR-10. {\bf b)} Maximum eigenvalue of the Hessian matrix at the end of training for \textbf{LB} with Ghost Batch Normalization, \textbf{LB} with Ghost Batch Normalization + diagonal Fisher and \textbf{SB}. Error bar is computed over 3 random seeds.}
\vspace{-0.2cm}
\end{figure}

We provide the details of this experiment. Computing maximum eigenvalue without any modification to the model gives inconsistent estimates even between different runs of the same training configuration. To make the maximum eigenvalue of the Hessian comparable over different training trajectories, the Hessian needs to be invariant under typical weight reparameterizations; for example, affine transformations~\citep{liao2018surprising}. To achieve this, we make the following modification to the trained model: (1) For the layers with batch normalization, we can just push the batch-norm layer parameters and running statistics into the layerwise parameter space so that the the layer is invariant under affine transformations; and (2) For the layers without batch normalization, reparameterization changes the prediction confidence while the prediction remains the same. Hence, we train a temperature parameter on the cross-entropy loss on the test data set, this encourages the model to make a calibrated prediction (prevent it from being over-confident). In Fig.~\ref{fig:spectral}, we give the error bar of the maximum eigenvalue of the Hessian over different runs, which indicates the modification gives a roughly consistent estimate. 

In Fig.~\ref{fig:spectral}, we give the error bar of the maximum eigenvalue of the Hessian over different runs for the regimes we experiment with. The central takeaway here is that across all models with which we experiment, we consistently find that the maximum eigenvalue of the Hessian for {\bf LB} with diagonal Fisher is lower than that of {\bf LB} and in some cases, comparable to {\bf SB}. 

\begin{table*}[t!]
\caption{Validation accuracy results on classification tasks for {\bf SB}, {\bf LB}, {\bf LB} + Fisher Trace, and {\bf LB} + diagonal Fisher. The results are averaged over 3 random seeds. All methods in each row are trained with the same number of epochs. While it is indeed not feasible to experiment {\bf LB} with full Fisher for all the models below, we note that this reaches roughly the same validation accuracy (93.22) as {\bf SB} in the case of ResNet44 (CIFAR-10).}
\label{tab:gen-table}
\begin{center}
\begin{small}
\scalebox{0.95}{
\begin{sc}
\begin{tabular}{lcc|cccr}
\toprule
Dataset & Model & SB & LB & LB+Fisher Trace & {\bf LB+Diag} \\
\midrule
MNIST & MLP &  98.10 & 97.95 & 98.08 & {\bf 98.10} \\
MNIST & LeNet &  99.10 & 98.88 & 99.02 & {\bf 99.11} \\
FASHION & LeNet & 91.15 & 88.89 & 90.29 & {\bf 90.79}  \\
CIFAR-10 & Alexnet & 87.80 & 86.42 & N/A & {\bf 87.61}  \\  
CIFAR-100 & Alexnet & 59.21 & 56.79 & N/A & {\bf 59.10}  \\  
CIFAR-10 & VGG16 & 93.25 & 91.81 & 92.91 & {\bf 93.19}  \\  
CIFAR-100 & VGG16 & 72.83 & 69.45 & 71.35 & {\bf 72.11}  \\
CIFAR-10 & ResNet44 & 93.42 & 91.93 & 92.33 & {\bf 92.88}  \\
CIFAR-100 & ResNet44x2 & 75.55 & 73.13 & 73.77 & {\bf 74.26} \\
\bottomrule
\end{tabular}
\end{sc}
}
\end{small}
\end{center}
\vskip -0.1in
\end{table*}

\vspace{-1mm}
\subsection{Generalization Gap} 

In this last part of our empirical studies, we apply our methodology to address the~\enquote{generalization gap} problem in stochastic optimization. Here, we experiment with four regimes: {\bf SB}, {\bf LB}, {\bf LB} with Fisher Trace and {\bf LB} with diagonal Fisher. All regimes are trained for the same number for epochs. {\bf LB} with Fisher Trace refers to {\bf LB} with the following noise  $\sqrt{\Tr(F(\theta_k))/d}\cdot\xi_k,\xi_k\sim\NN(0,I_d)$ injected at every iteration. The purpose of this is to provide an experimental comparison which adds isotropic noise with the same variance as {\bf LB} with diagonal Fisher.  

We also point out that we do not experiment with {\bf LB} with full Fisher due to its exceeding long training time. This can be seen from Fig.~\ref{fig:train_loss} where for ResNet44 trained on CIFAR-10, {\bf LB} with full Fisher does not achieve good convergence even after 8000 parameter updates. There is typically no optimal learning-rate scaling rule for large-batch training across different models and datasets~\citep{shallue2018measuring}; hence, we tune the learning rate schedule to obtain optimal results for each method.

The final validation accuracy numbers are reported in Table~\ref{tab:gen-table}. While it is true that using {\bf LB} with diagonal Fisher cannot completely close the~\enquote{generalization gap} in some cases, it yields definite improvements over {\bf SB} within an epoch-training budget. Such a training regime typically favors small-batch training as they perform more parameter updates~\citep{shallue2018measuring}. This highlights that our approach is a data-efficient way to improve generalization for large-batch training.

In addition, we experimented with other regimes such as injecting multiplicative Gaussian noise with constant covariance as in~\citet{hoffer2017train} and replacing diagonal Fisher with the block-diagonal Kronecker-Factored Approximate Curvature (K-FAC)~\citep{martens2015optimizing}~\footnote{This corresponds to the matrix-variate Gaussian noise in~\citet{zhang2018noisy}}. We delegate these results to Section~\ref{sec:extra-results} of Appendix.

\vspace{-1mm}
\section{Related Works}
\label{sec:related works}

\textbf{Variance and Optimization.} In the context of large-scale learning, stochastic algorithms are very popular compared to full-batch methods due to lower computational overhead~\citep{bottou2018optimization, Bottou91stochasticgradient}. The tradeoff is that stochastic algorithms exhibit slower convergence asymptotically due to the inherent noise present in their gradients~\citep{moulines2011non, bottou2018optimization,wen2018flipout}. For smooth and strongly-convex functions, variance reduction is a common technique to improve convergence rates~\citep{JZ:13, DBL:14}. 

In contrast, for non-convex optimization, increasing the variance by adding noise is often times beneficial. Unlike the convex setting, the loss surface is much more complicated and there is an abundance of global minima~\citep{Choromanska2015}. Adding noise can significantly improve training since it enables the dynamics to escape saddle points or shallow local minima~\citep{ge2015escaping,jin2017escape}. More specifically for deep learning, injecting annealed gradient noise has been shown to speed up training of very deep neural networks~\citep{Neelakantan:15}. 

\textbf{Variance and Generalization.} The inherent noise in stochastic optimization methods is also conducive to generalization performance. There are vast bodies of literature devoted to this in deep learning; for example, scaling the learning rate or batch-size~\citep{SL:17, goyal2017accurate, hoffer2017train} to augment gradient noise in order to encourage better generalization. More direct approaches of studying the covariance structure of mini-batch gradients have also been explored~\citep{jastrzkebski2017three, xing2018walk, ZWYM:18, li2015stochastic}. A closely-related approach to ours is the Stochastic Gradient Langevin Dynamics (SGLD)~\citep{Gelfand:1992,welling2011bayesian}; a modification of SGD where an annealed isotropic Gaussian noise is injected to the gradients. The recent systematic empirical study of~\citet{shallue2018measuring} demonstrates that links between optimization, generalization, and the choice of batch-size in SGD is extremely complex. It underscores the necessity of a more foundational understanding of the interaction between batch-sizes, model architectures, and other optimization metaparameters.

\vspace{-1.5mm}
\section{Conclusion}
\label{sec:conclusion}

In this paper, we explored using covariance noise in designing optimization algorithms for deep neural networks that could potentially exhibit ideal learning behavior. 
We proposed to add diagonal Fisher noise to large-batch gradient updates. Our empirical studies showed that this yield significant improvements in generalization while retaining desirable convergence performance. Furthermore, we demonstrated that the structure of the noise covariance matrix encodes much more information about optimization than the variance of gradients. An immediate question which arises is to better understand how the mathematical structure of the noise covariance matrix ties to generalization. For example, in the special cases of diagonal Fisher and full Fisher, our experiments appear to indicate that the generalization performance is somewhat comparable. This seems to suggest that the diagonal elements contribute much more to the generalization performance than the off-diagonal elements. It is an interesting theoretical question as to why this may be the case.

\bibliography{refs}
\bibliographystyle{plainnat}

\appendix
\section{SUPPLEMENTARY MATERIAL}

\subsection{Proof of Theorem~\ref{convergence analysis theorem}} \label{sec:proof of convergence theorem}

The proof of this theorem follows the spirit of~\citet{bottou2018optimization}. The algorithm 
\begin{equation} \label{eq:constant convariance algorithm}
\theta_{k+1} = \theta_k - \alpha_k \nabla\LL(\theta_k) + \alpha_k C\xi_{k+1},\ \xi_{k+1}\sim\NN(0,I_d).
\end{equation}
falls into the Robbins-Monro setting where the true gradient is perturbed by random noise. This perturbation can be considered as a martingale
difference in the sense that
\[
\EE[C \xi_{k+1}  | \F_k] =  0
\]
where $(\F_k)_{k \in \N}$ is a increasing filtration generated by the
sequence of parameters $(\theta_k)_{k \in \N}$. When the step size is constant $\alpha_k=\alpha$ for all $k$, it corresponds to the Euler discretization of a gradient flow with random perturbation. 
We begin the proof by considering the equality, 
\begin{align*}
\LL(\theta_{k+1}) & = \LL(\theta_k) + \langle\LL(\theta_k),\theta_{k+1}-\theta_k\rangle  \\
& + \frac{1}{2}(\theta_{k+1}-\theta_k)^{\top}\nabla^2\LL(\theta_k)(\theta_{k+1}-\theta_k).
\end{align*}
Using the fact that $\nabla\LL(\theta_k)=A\theta_k$, $\nabla^2\LL(\theta_k)=A$, and from the definition of $\theta_{k+1}$, we can rewrite the above equation as
\begin{align*}
\LL(\theta_{k+1}) & = \LL(\theta_k) + \langle A\theta_k,-\alpha_k A\theta_k+\alpha_k C\xi_{k+1}\rangle \\
& + \frac{1}{2}\norm{\alpha_k A\theta_k-\alpha_k C\xi_{k+1}}_A^2.
\end{align*}
Now, taking the conditional expectation $\EE[\cdot|\F_k]$ on both sides of the equality, we obtain by independence of the noise $\xi_{k+1}$ to $\F_k$
\begin{equation} \label{eq:intermediate bound term}
\begin{aligned}
\EE[\LL(\theta_{k+1})|\F_k] & = \LL(\theta_k)-\alpha_k\norm{A\theta_k}_2^2 + 
\frac{\alpha_k^2}{2}\norm{A\theta_k}_A^2 \\
& +\frac{\alpha_k^2}{2}\EE[\norm{C\xi_{k+1}}_A^2]
\end{aligned}
\end{equation}
A simple computation shows
\begin{equation} \label{eq:computation of trace}
\begin{aligned}
\EE[\norm{C\xi_{k+1}}_A^2] & = \EE[(C\xi_{k+1})^{\top}A(C\xi_{k+1})] \\
& = \EE[\xi_{k+1}^{\top}C^{\top}AC\xi_{k+1}] \\
& = \Tr(C^{\top}AC)
\end{aligned}
\end{equation}
Moreover, we have
\begin{equation} \label{eq:norm/max-eigenvalue inequality}
\norm{A\theta_k}_A^2\leq\lambda_{\max}\norm{A\theta_k}_2^2,
\end{equation}
where $\lambda_{\max}$ denotes maximum eigenvalue of $A$ (and $\lambda_{\min}$ correspondingly denotes minimum eigenvalues of $A$). This comes from the fact that for any symmetric, positive-definite matrix $B$, 
\begin{equation} \label{eq:matrix norm-max eigenvalue inequality}
\norm{x}_B^2\leq\lambda_{\max, B}\norm{x}_2^2,
\end{equation}
where $\lambda_{\max, B}$ denotes maximum eigenvalue of $B$.
Using the results in Eqns.~\ref{eq:computation of trace} and~\ref{eq:norm/max-eigenvalue inequality} as well as the assumption on the step-size schedule for all $k$: $\alpha_k < \alpha_0 < \frac{1}{\lambda_{\max}}$, we rewrite Eqn.~\ref{eq:intermediate bound term} as
\begin{equation} \label{eq:bound in 2-norm}
\begin{aligned}
\EE[\LL(\theta_{k+1})|\F_k] & \leq \LL(\theta_k) + \left(\frac{\alpha_k}{2}\lambda_{\max}-1\right)\alpha_k\norm{A\theta_k}_2^2 \\
& +\frac{\alpha_k^2}{2}\Tr(C^{\top}AC) \\
& \leq \LL(\theta_k) - \frac{\alpha_k}{2}\norm{A\theta_k}_2^2 + \frac{\alpha_k^2}{2}\Tr(C^{\top}AC).
\end{aligned}
\end{equation}
Now, taking $x=A\theta_k$ and $B=A^{-1}$ in Eqn.~\ref{eq:matrix norm-max eigenvalue inequality}, we have
\[
\norm{A\theta_k}_{A^{-1}}^2\leq\lambda_{\max, A^{-1}}\norm{A\theta_k}_2^2
\]
Note that $\lambda_{\max,A^{-1}}=\lambda_{\min}$ and simplifying the above,
\begin{align*}
\norm{A\theta_k}_2^2 & \geq\lambda_{\min}\norm{A\theta_k}_{A^{-1}}^2 \\
& = \lambda_{\min}(\theta_k^{\top}A)A^{-1}(A\theta_k) \\
& = \lambda_{\min}\norm{\theta_k}_A^2 \\
& = 2\lambda_{\min}\LL(\theta_k).
\end{align*}
Using the above inequality in Eqn.~\ref{eq:bound in 2-norm} and then taking expectation yields
\[
\EE[\LL(\theta_{k+1})] \leq (1-\alpha_k\lambda_{\min})\EE[\LL(\theta_k)] + \frac{\alpha_k^2}{2}\Tr(C^{\top}AC).
\]
We proceed by induction to prove the final result. By definition of $\nu$, the result is obvious for $k=0$. For the inductive step, suppose that the induction hypothesis 
holds for $k$, i.e.,
\begin{equation*}
\alpha_k = \frac{2}{(k+\gamma)\lambda_{\min}}, \quad \EE[\LL(\theta_k)] \leq \frac{\nu}{k+\gamma}.
\end{equation*}
We prove the $k+1$ case.
\begin{align*}
\EE[\LL(\theta_{k+1})] &\leq \left(1- \frac{2 }{k+\gamma }\right)\frac{\nu}{k+ \gamma}
\\ 
& + \frac{2 }{(k+\gamma)^2 \lambda^2_{\min}}\Tr(C^{\top}AC)\\
&\leq \frac{\nu}{(k+\gamma+1)}
\end{align*}
This comes from the definition of $\nu$ and also the inequality $ (k+ \gamma -1)(k+\gamma + 1) \leq (k+\gamma)^2$. This concludes the proof.
\qed

\subsection{Relationship Between Noise Covariance Structures and Generalization} \label{sec:covariance-generalization}

As in the previous section, we work entirely in the convex quadratic setting. In this case, Eqn.~\ref{eq:naive LB update} becomes 
\begin{equation} \label{eq:constant covariance algorithm}
\theta_{k+1} = \theta_k-\alpha_k\nabla\LL(\theta_k)+\alpha_kC\xi_{k},\ \xi_{k}\sim\NN(0,I_d).
\end{equation}
Our aim in this section is to provide some theoretical discussions on how the choice of covariance structure $C$  influences the generalization behavior.

\textbf{Uniform stability.} Uniform stability~\citep{bousquet2002stability} is one of the most common techniques used in statistical learning theory to study generalization of a learning algorithm. Intuitively speaking, uniform stability measures how sensitive an algorithm is to perturbations of the sampling data. The more stable an algorithm is, the better its generalization will be. 
Recently, the uniform stability has been investigated for Stochastic Gradient methods~\citep{Hardt:15} and also for Stochastic Gradient Langevin Dynamics (SGLD)~\citep{mou2017generalization, Raginsky17}. We present the precise definition.

\begin{Definition}[Uniform stability]
A randomized algorithm $\AAA$ is $\epsilon$-stable if for all data sets $\lS$ and $\lS'$ where $\lS$ and $\lS'$ differ in at most one sample, we have
\[
\sup_{(x,y)}|\EE_{\AAA}[\LL(\theta_{\lS})-\LL(\theta_{\lS'})]|\leq\epsilon,
\]
where $\LL(\theta_{\lS})$ and $\LL(\theta_{\lS'})$ highlight the dependence of parameters on sampling datasets. The supremum is taken over input-target pairs $(x,y)$ belonging to the sample domain.
\end{Definition}

The following theorem from~\citet{bousquet2002stability} shows that uniform stability implies generalization.

\begin{Theorem}[Generalization in expectation]
Let $\AAA$ be a randomized algorithm which is $\epsilon$-uniformly stable, then 
\[
|\EE_{\AAA}[\EEE_{\mathrm{gen}}]|\leq\epsilon,
\]
where $\EEE_{\mathrm{gen}}$ is the expected generalization error as defined in Eqn. 1 of Section 2.
\end{Theorem}

{\bf Continuous-time dynamics.} We like to use the uniform stability framework to analyze generalization behavior of Eqn.~\ref{eq:constant convariance algorithm}. To do this, we borrow ideas from the recent work of~\citet{mou2017generalization} which give uniform stability bounds for Stochastic Gradient Langevin Dynamics (SGLD) in non-convex learning. While the authors in that work give uniform stability bounds in both the discrete-time and continuous-time setting, we work with the continuous setting since this conveys relevant ideas while minimizing technical complications. The key takeaway from~\citet{mou2017generalization} is that uniform stability of SGLD may be bounded in the following way
\begin{equation} \label{eq:SGLD uniform stability}
\epsilon_{\mathrm{SGLD}}\leq\sup_{\lS,\lS'} \sqrt{H^2(\pi_t,\pi_t')}.
\end{equation}
Here, $\pi_t$ and $\pi_t'$ are the distributions on parameters $\theta$ trained on the datasets $\lS$ and $\lS'$. The $H^2$ refers to the Hellinger distance.

We now proceed to mirror the approach of~\citet{mou2017generalization} for Eqn.~\ref{eq:constant convariance algorithm}. Our usage of stochastic differential equations will be very soft but we refer to reader to~\citet{gardiner2009stochastic,pavliotis2014stochastic} for necessary background. For the two datasets $\lS$ and $\lS'$, the continuous-time analogue of Eqn.~\ref{eq:constant convariance algorithm} are Ornstein-Uhlenbeck processes~\citep{uhlenbeck1930theory}:
\begin{align*}
d\theta_{\lS}(t) = -A_{\lS}\theta_{\lS}(t)dt + \sqrt{\alpha} C_{\lS} dW(t) \\
d\theta_{\lS'}(t) = -A_{{\lS}'}\theta_{{\lS}'}(t)dt + \sqrt{\alpha} C_{{\lS}'} dW(t).
\end{align*}
The solution is given by
\[
\theta_{\lS}(t) = e^{-A_{\lS} t}\theta_{\lS}(0) + \sqrt{\alpha}\int^t_0  e^{-A_{\lS} (t-u)}C_{\lS}dW(u),
\]
In fact, this yields the Gaussian distribution
\[
\theta_{\lS}(t)\sim\NN(\mu_{\lS}(t),\Sigma_{\lS}(t)),
\]
where 
\[
\mu_{\lS}(t) = e^{-A_{\lS} t}\theta_{\lS}(0) 
\]
and $\Sigma_{\lS}(t)$ satisfies the Ricatti equation,
\[
\frac{d}{dt}\Sigma_{\lS}(t) = -(A_{\lS}\Sigma_{\lS}(t) + \Sigma_{\lS}(t)A_{\lS})+ \alpha C_{\lS}C_{\lS}^{\top}. 
\]
Observe that $A_{\lS}$ is symmetric and positive-definite which means that it admits a diagonalization $A_{\lS}=P_{\lS}D_{\lS}P_{\lS}^{-1}$. Solving the equation for the covariance matrix gives
\begin{align} \label{eq:covariance of parameter distributions}
\Sigma_{\lS}(t) &= \alpha P_{\lS} \left(\int^t_0 e^{-D_{\lS}(t-u)}P_{\lS}^{-1}C_{\lS}C_{\lS}^{\top}P_{\lS}e^{-D_{\lS}(t-u)}du \right)P_{\lS}^{-1}.
\end{align}
We are in the position to directly apply the framework of~\citep{mou2017generalization}. Choosing $\pi_t$ and $\pi_{t'}$ in Eqn.~\ref{eq:SGLD uniform stability} to be the Gaussians $\NN(\mu_{\lS}(t),\Sigma_{\lS}(t))$ and $\NN(\mu_{\lS'}(t),\Sigma_{\lS'}(t))$ respectively, we obtain a uniform stability bound for Eqn.~\ref{eq:constant convariance algorithm}. We compute the right-hand side of the bound to obtain insights on generalization. Using the standard formula for Hellinger distance between two Gaussians, we have
\begin{equation} \label{eq:Hellinger distance expanded}
H^2(\pi_t,\pi_{t}') = 1-\frac{\det(\Sigma_{\lS})^{\frac{1}{4}}\det(\Sigma_{\lS'})^{\frac{1}{4}}}{\det(\frac{\Sigma_{\lS}+\Sigma_{\lS'}}{2})^{\frac{1}{2}}}\Lambda_{\lS,\lS'}
\end{equation}
where $\Lambda_{\lS,\lS'}$ is
\[
\exp\left\{-\frac{1}{8}(\mu_{\lS}-\mu_{\lS'})^{\top}\left(\frac{\Sigma_{\lS}+\Sigma_{\lS'}}{2}\right)^{-1}(\mu_{\lS}-\mu_{\lS'})\right\}.
\]

{\bf Choosing the noise covariance.} From Eqn.~\ref{eq:Hellinger distance expanded} above, it is evident that to ensure good generalization error for Eqn.~\ref{eq:constant convariance algorithm}, we want to choose a covariance $C_{\lS}$ such that the Hellinger distance $H^2$ is minimized. Since we are working within the uniform stability framework, a good choice of $C_{\lS}$ should be one where Eqn.~\ref{eq:constant convariance algorithm} becomes less data-dependent. This is intuitive after all -- the less data-dependent an algorithm is; the better suited it should be for generalization.

We study Eqn.~\ref{eq:Hellinger distance expanded}. Note that as time $t\to\infty$, the exponential term goes to 1. Hence, we focus our attention on the ratio of the determinants. 
Suppose that we choose $C_{\lS}=\sqrt{A_{\lS}}$ and note that $A_{\lS}$ is the Fisher in this convex quadratic example. Simplifying the determinant of $\Sigma_{\lS}(t)$ in this case, \begin{align*}
\det(\Sigma_{\lS}(t)) 
& = \left( \frac{\alpha}{2}\right)^{d}\det(I_d-e^{-2D_{\lS}t})
\end{align*}
Suppose that we choose $C=I_d$. Proceeding analogously,
\begin{align*}
\det(\Sigma_{\lS}(t)) 
& = \left(\frac{\alpha}{2} \right)^d \frac{\det(I_d-e^{-2D_{\lS}t})}{\det(D_{\lS})}
\end{align*}
We can think of choosing $C=I_d$ or  $C=\sqrt{A}$ to be extreme cases and it is interesting to observe that the Hellinger distance is more sensitive to dataset perturbation when $C=I_d$.
Our proposed method of this paper was to choose $C=\sqrt{\diag{A}}$ and our experiments seem to suggest that choosing the square-root of diagonal captures much of the generalization behavior of full Fisher. Understanding precisely why this is the case poses an interesting research direction to pursue in the future.

A simple scaling argument also highlights the importance of the trade-off between optimization and generalization. Consider $C_{\lambda} = \lambda C$.
Then Theorem~\ref{convergence analysis theorem} suggests to take $\lambda$ small to reduce the variance and improve convergence.
However, in that case $\Sigma_{\lambda} = \lambda^2 \Sigma$ where $\Sigma$ is given by the Eqn.~\ref{eq:covariance of parameter distributions} for $C$ and
\begin{equation*} 
H^2(\pi_t,\pi_{t}') = 1-\frac{\det(\Sigma_{\lS})^{\frac{1}{4}}\det(\Sigma_{\lS'})^{\frac{1}{4}}}{\det(\frac{\Sigma_{\lS}+\Sigma_{\lS'}}{2})^{\frac{1}{2}}}\Lambda_{\lS,\lS',\lambda},
\end{equation*}
where $\Lambda_{\lS,\lS',\lambda}$ is
\[
\exp\left\{-\frac{1}{8\lambda^2}(\mu_{\lS}-\mu_{\lS'})^{\top}\left(\frac{\Sigma_{\lS}+\Sigma_{\lS'}}{2}\right)^{-1}(\mu_{\lS}-\mu_{\lS'})\right\}.
\]
The Hellinger distance gets close to one in the limit of small $\lambda$ (which intuitively corresponds to the large batch situation).

\subsection{Fisher Information Matrix for Deep Neural Networks} \label{sec:Fisher for different architectures}

In this section, we give a formal description of the Fisher information matrix for both feed-forward networks and convolutional networks. In addition, we give the diagonal expression for both networks. Note that these expressions are valid for both the empirical and the exact Fisher; in the empirical case, the expectation will be taken over the empirical data distribution whereas in the exact case, the expectation will be taken over the predictive distribution for targets $y$.

\subsection{Feed-forward networks}

Consider a feed-forward network with $L$ layers. At each layer $i\in\{1,\dots,L\}$, the network computation is given by
\begin{align*}
z_i & = W_i a_{i-1} \\
a_i & = \phi_i(z_i),
\end{align*}
where $a_{i-1}$ is an activation vector, $z_i$ is a pre-activation vector, $W_i$ is the weight matrix, and $\phi_i:\RR\to\RR$ is a nonlinear activation function applied coordinate-wise. Let $w$ be the parameter vector of network obtained by vectorizing and then concatenating all the weight matrices $W_i$,
\[
w=[\ovec(W_1)^{\top}\ \ovec(W_2)^{\top}\ \dots\ \ovec(W_L)^{\top}]^{\top}.
\]
Furthermore, let $\DD v=\nabla_{v}\log p(y|x,w)$ denote the log-likelihood gradient. Using backpropagation, we have a decomposition of the log-likelihood gradient $\DD W_i$ into the outer product:
\[
\DD W_i = g_i a_{i-1}^{\top},
\]
where $g_i=\DD z_i$ are pre-activation derivatives. The Fisher matrix $F(w)$ of this feed-forward network is a $L\times L$ matrix where each $(i,j)$ block is given by
\begin{equation} \label{eq:i,j blocks of Fisher}
F_{i,j}(w) = \EE[\ovec(\DD W_i)\ovec(\DD W_j)^{\top}] = \EE[a_{i-1}a_{j-1}^{\top}\otimes g_ig_j^{\top}].
\end{equation}

{\bf Diagonal version.} We give an expression for the diagonal of $F_{i,i}(w)$ here. The diagonal of $F(w)$ follows immediately afterwards. Let $a_{i-1}^2$ and $g_i^2$ be the element-wise product of $a_{i-1}$ and $g_i$ respectively. Then, in vectorized form, 
\[
\diag{F_{i,i}(w)} = \EE[\ovec((a_{i-1}^2)(g_i^2)^{\top})],
\]
where $(a_{i-1}^2)(g_i^2)^{\top}$ is the outer product of $a_{i-1}^2$ and $g_i^2$. 

\subsection{Convolutional networks}

In order to write down the Fisher matrix for convolutional networks, it suffices to only consider convolution layers as the pooling and response normalization layers typically do not contain (many) trainable weights. We focus our analysis on a single layer. Much of the presentation here follows~\citep{grosse2016kronecker,NG-invariances}.

A convolution layer $l$ takes as input a layer of activations $a_{j,t}$ where $j\in\{1,\dots,J\}$ indexes the input map and $t\in\TT$ indexes the spatial location. $\TT$ here denotes the set of spatial locations, which we typically take to be a 2D-grid. We assume that the convolution here is performed with a stide of 1 and padding equal to the kernel radius $R$, so that the set of spatial locations is shared between the input and output feature maps. This layer is parameterized by a set of weights $w_{i,j,\delta}$, where $i\in\{1,\dots,I\}$ indexes the output map and $\delta\in\Delta$ indexes the spatial offset. The numbers of spatial locations and spatial offsets are denoted by $|\TT|$ and $|\Delta|$ respectively. The computation of the convolution layer is given by
\begin{equation} \label{eq:convolutional computation}
z_{i,t} = \sum_{\delta\in\Delta} w_{i,j,\delta}a_{j,t+\delta}. 
\end{equation}
The pre-activations $z_{i,t}$ are then passed through a nonlinear activation function $\phi_l$. The log-likelihood derivatives of the weights are computed through backpropagation:
\[
\DD w_{i,j,\delta} = \sum_{t\in\TT}a_{j,t+\delta}\DD z_{i,t}.
\]
Then, the Fisher matrix here is 
\[
\EE\left[\left(\sum_{t\in\TT}a_{j,t+\delta}\DD z_{i,t}\right)\left(\sum_{t'\in\TT}a_{j',t'+\delta'}\DD z_{i',t'}\right)\right]
\]

\begin{table*}[t]
\caption{Validation accuracy results on classification tasks using BatchChange, Multiplicative, K-FAC and Fisher Trace. Results are averaged over 3 random seeds. For the reader’s convenience, we report again the result of Diag-F.}
\label{tab:extra-gen-table}
\begin{center}
\begin{small}
\begin{sc}
\begin{tabular}{lcccccccr}
\toprule
Dataset & Model & SB & BatchChange & Multiplicative & K-FAC & Fisher Trace & Diag-F \\
\midrule
CIFAR-10 & VGG16 & 93.25 & 93.18 & 90.98 & 93.06 & 92.91 & 93.19 \\  
CIFAR-100 & VGG16 & 72.83 & 72.44 & 68.77 & 71.86 & 71.35 & 72.11  \\  
CIFAR-10 & ResNet44 & 93.42 & 93.02 & 91.28 & 92.81 & 92.33 & 92.88  \\  
CIFAR-100 & ResNet44x2 & 75.55 & 75.16 & 71.98 & 73.84 & 73.77 & 74.26 \\

\bottomrule
\end{tabular}
\end{sc}
\end{small}
\end{center}
\vskip -0.1in
\end{table*}

{\bf Diagonal version.} To give the diagonal version, it will be convenient for us to express the computation of the convolution layer in matrix notation. First, we represent the activations $a_{j,t}$ as a $J\times|\TT|$ matrix $A_{l-1}$, the pre-activations $z_{i,t}$ as a $I\times|\TT|$ matrix $Z_l$, and the weights $w_{i,j,\delta}$ as a $I\times J|\Delta|$ matrix $W_l$. Furthermore, by extracting the patches surrounding each spatial location $t\in\TT$ and flattening these patches into column vectors, we can form a $J|\Delta|\times|\TT|$ matrix $A_{l-1}^{\exp}$ which we call the expanded activations. Then, the computation is Eqn.~\ref{eq:convolutional computation} can be reformulated as the matrix multiplication
\[
Z_l = W_l A_{l-1}^{\exp}. 
\]
Readers familiar with convolutional networks can immediately see that this is the Conv2D operation. 

At a specific spatial location $t\in\TT$, consider the $J|\Delta|$-dimensional column vectors of $A_{l-1}^{\exp}$ and $I$-dimensional column vectors of $Z_l$. Denote these by $a_{l-1}^{(:,t)}$ and $z_l^{(t)}$ respectively. The matrix $W_l$ maps $a_{l-1}^{(:,t)}$ to $z_l^{(t)}$. In this case, we find ourselves in the exact same setting as the feed-forward case given earlier. The diagonal is simply
\[
\EE\left[\ovec\left((a_{l-1}^{(:,t)})^2(\DD z_l^{(t)})^2\right)\right]
\]

\subsection{Kronecker-Factored Approximate Curvature (K-FAC)}

Later in Section~\ref{sec:supplementary experiments}, we will compare the diagonal approximation of the Fisher matrix to the Kronecker-factored approximate curvature (K-FAC)~\citep{martens2015optimizing} approximation of the Fisher matrix. We give a brief overview of the K-FAC approximation in the case of feed-forward networks.

Recall that the Fisher matrix for a feed-forward network is a $L\times L$ matrix where each of the $(i,j)$ blocks are given by Eqn.~\ref{eq:i,j blocks of Fisher}. Consider the diagonal $(i,i)$ blocks. If we approximate the activations $a_{i-1}$ and pre-activation derivatives $g_i$ as statistically independent, we have
\begin{align*}
F_{i,i}(w) & =\EE[\ovec(\DD W_i)\ovec(\DD W_i)^{\top}] \\
& = \EE[a_{i-1}a_{i-1}^{\top}\otimes g_i g_i^{\top}] \\
& \approx\EE[a_{i-1}a_{i-1}^{\top}]\otimes\EE[g_ig_i^{\top}].
\end{align*}
Let $A_{i-1}=\EE[a_{i-1}a_{i-1}^{\top}]$ and $G_i=\EE[g_ig_i^{\top}]$. The K-FAC approximation $\hat{F}$ of the Fisher matrix $F$ is
\[
\hat{F} = \left[\begin{array}{cccc}
	A_{0}\otimes G_{1} &  &  & 0\\
	& A_{1}\otimes G_{2}\\
	&  & \ddots\\
	0 &  &  & A_{L-1}\otimes G_{L}
\end{array}\right].
\]
The K-FAC approximation of the Fisher matrix can be summarized in the following way: (1) keep only the diagonal blocks corresponding to individual layers, and (2) make the probabilistic modeling assumption where the activations and pre-activation derivatives are statistically independent. 

\subsection{Supplementary Experiments Details} \label{sec:supplementary experiments}

{\bf Learning rate}: We tuned the learning rate schedule for each method in Table~\ref{tab:gen-table} of Section~\ref{sec:moreexperiments} to obtain best performance. As a result, for both {\bf LB} and {\bf LB} with diagonal Fisher method, we need to scale up the learning rate and use the linear warmup strategy in the first 10 epochs. For {\bf LB}, the optimal learning rate on CIFAR-10 and CIFAR-100 with ResNet44 is 3.2 while is 1.6 for {\bf LB} with diagonal Fisher. With VGG16 network on CIFAR-10 and CIFAR-100, the optimal learning rates are 1.6 for both methods. We decay the learning rate by 0.1 at the epoch of 100, 150 for all above methods.

{\bf Noise Termination}: For all training regimes involving noise injection, we found terminating the noise at a quarter of the training trajectory and using standard {\bf LB} for the remainder of training achieves the best performance. This finding is consistent to the result of BatchChange in Table~\ref{tab:extra-gen-table}, which suggests that noise only helps generalization in the beginning of the training.

\subsection{Validation Accuracy Results} \label{sec:extra-results}

We provide additional validation accuracy results to complement Table~\ref{tab:gen-table} of Section~\ref{sec:moreexperiments}. The additional regimes are:
\begin{itemize}
    \item \textbf{BatchChange:} Here, we use {\bf SB} for the first 50 epochs and then use {\bf LB} for the remainder. This experimental setup was inspired by~\citet{smith2017don}.
    \item \textbf{Multiplicative:} Here, we multiply the gradients with a Gaussian noise with constant diagonal covariance structure. This experimental setup was inspired by~\citet{hoffer2017train}.
    \item \textbf{K-FAC:} Instead of choosing diagonal Fisher as the noise covariance structure, we use the block-diagonal approximation of Fisher given by K-FAC instead
    \item \textbf{Fisher Trace:} Instead of choosing diagonal Fisher as the noise covariance structure, we use square-root of the trace of Fisher $\sqrt{\Tr(F(\theta_k))}$ instead
\end{itemize}

The results are reported in Table~\ref{tab:extra-gen-table} above. 

\subsection{Sampling Full Fisher Noise} \label{sec:full fisher noise}

\textbf{Sampling True Fisher Random Vector.} We describe a method to sample a random vector with Fisher covariance efficiently. We obtain prediction $f(x,\theta)$ by a forward-pass. If we randomly draw labels from the model's predictive distribution and obtain back-propagated gradients $\nabla_{\theta} \LL$, then we have $\mathrm{Cov}(\nabla_{\theta} \LL, \nabla_{\theta} \LL)=\EE_x[J_f^{\top}H_{\LL}J_f]$, which is the exact true Fisher~\citep{martens2014new}. Here, $J_f$ is the Jacobian of outputs with respect to parameters and $H_{\LL}$ is the Hessian of the loss function with respect to the outputs.

\textbf{Sampling Empirical Fisher Random Vector.}
Let $M$ be the size of the mini-batch and from the $M$-forward passes we obtain the back-propagated gradients $\nabla l_1, \dots , \nabla l_M$ for each data-point. Consider independent random variables $\sigma_1,\dots,\sigma_M$ drawn from Rademacher distribution, i.e.,
$P(\sigma_i = 1) = P(\sigma_i = -1)=\frac{1}{2}$. Then, the mean
$\EE_{\sigma}[\sum_{i=1}^M \sigma_i\nabla l_i] = 0$. The covariance is empirical Fisher.
\end{document}